\pdfoutput=1

\documentclass[letterpaper, 10 pt, conference]{ieeeconf} 

\IEEEoverridecommandlockouts  

\usepackage{todonotes}

\usepackage{verbatim} 
\usepackage{subfigure}
\usepackage{tabulary}
\usepackage{tabularx}
\usepackage{balance}
\usepackage{mathtools}
\usepackage{textcomp}
\usepackage{multirow}
\usepackage{amssymb}  
\usepackage{url}
\definecolor{CommentRed}{rgb}{0.7,0,0}
\definecolor{CommentBlue}{rgb}{0,0,0.7}
\definecolor{CommentDG}{rgb}{0,0.6,0}

\usepackage{color, colortbl}
\definecolor{LightCyan}{rgb}{0.88,1,1}

\usepackage{tikz, pgfplots, fix-cm}
\usetikzlibrary{plotmarks}

\title{\LARGE \bf Fruit Quantity and Quality Estimation using a Robotic Vision System}

\author{$\text{Michael Halstead}^{*,\dagger}$, $\text{Christopher McCool}^{*,\dagger}$, {$\text{Simon Denman}^{*}$},  {$\text{Tristan Perez}^{*}$} and {$\text{Clinton Fookes}^{*}$} 
\thanks{* Science and Engineering Faculty, Queensland University of Technology, Brisbane, Australia.
\texttt{\{m.halstead, c.mccool\}@qut.edu.au} \newline
\indent $\dagger$ These authors contributed equally to this work.
}
}%

\begin{document}

\maketitle

\thispagestyle{empty}
\pagestyle{empty}


\begin{abstract}
Accurate localisation of crop remains highly challenging in unstructured environments such as farms.
Many of the developed systems still rely on the use of hand selected features for crop identification and often neglect the estimation of crop quantity and quality, which is key to assigning labor during farming processes.
To alleviate these limitations we present a robotic vision system that can accurately estimate the quantity and quality of sweet pepper (\textit{Capsicum annuum L}), a key horticultural crop. 
This system consists of three parts: detection, quality estimation, and tracking. 
Efficient detection is achieved using the FasterRCNN framework. 
Quality is then estimated in the same framework by learning a parallel layer which we show experimentally results in superior performance than treating quality as extra classes in the traditional FasterRCNN framework.
Evaluation of these two techniques outlines the improved performance of the parallel layer, where we achieve an $F_1$ score of 77.3 for the parallel technique yet only 72.5 for the best scoring (red) of the multi-class implementation.
To track the crop we present a tracking via detection approach, which uses the FasterRCNN with parallel layers, that is also a vision-only solution.
This approach is cheap to implement as it only requires a camera and in experiments across 2 days we show that our proposed system can accurately estimate the number of sweet pepper present, within 4.1\% of the ground truth.
\end{abstract}

\section{Introduction}

\begin{figure}[!t]
  \begin{center}
  {\includegraphics[width=0.8\columnwidth]{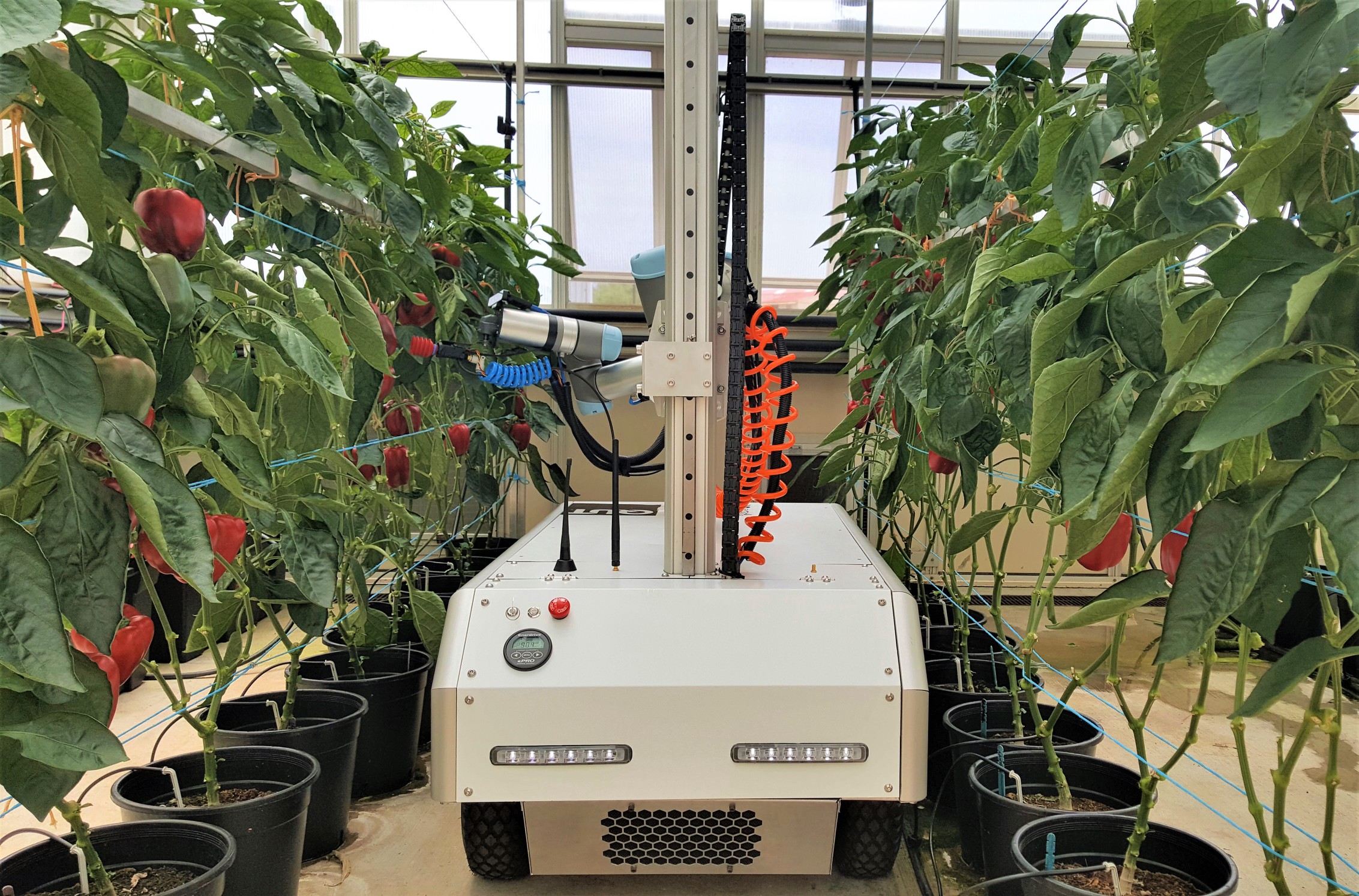}} \\
  \vspace{0.1cm}
  {\includegraphics[width=0.8\columnwidth]{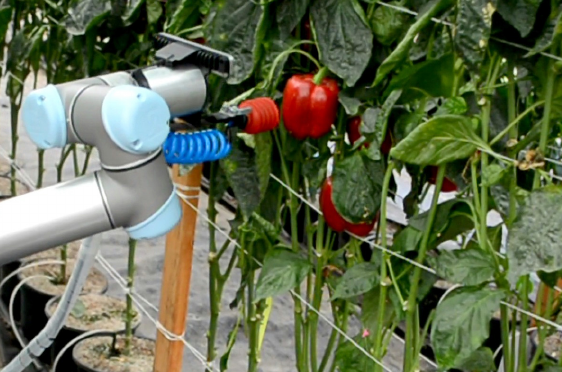}} 
  \end{center}
  \vspace{-15pt}
    \caption{
    Top is an image of Harvey in a protected cropping environment and bottom is a close up view of Harvey as it harvests fruit (sweet pepper).}
    \label{fig:FrontImage}
    \vspace{-20pt}
\end{figure}

Agricultural robotics are rapidly gaining interest as shown by the advent of weed management robots such as AgBot~II~\cite{Bawden17_1} and harvesting platforms such as Harvey~\cite{Lehnert17_1}, see Figure~\ref{fig:FrontImage}.
Robotic vision algorithms that allow these platforms to understand the diverse environments in which they operate are key for the operation of these robots. 
Harvesting robots, such as Harvey, detect and segment fruit in complicated scenes which often include high levels of occlusion.
Furthermore, the foliage in the scene can share similar colours to the fruit being detected (green on green).
Such problems, require advanced robotic vision techniques to deal with these challenging and unstructured environments.

Several researchers have explored the issue of crop detection, predominantly using traditional approaches.
Nuske et al.~\cite{Nuske_2011_6891} proposed a grape detection system based on a radial symmetry transform and used the predicted number of grapes to then accurately estimate the yield.
Conditional random fields were used by Hung et al.~\cite{Hung:2013aa} to perform almond segmentation and then by McCool et al.~\cite{McCool16_1:conference} to perform sweet pepper segmentation and detection.
All these approaches relied on hand-crafted features to perform crop detection or segmentation.

More recently, Sa et al.~\cite{Sa16_1} proposed the DeepFruits approach which employs a deep learning approach to accurately detect fruit.
Sa et al. explored the use of the Faster RCNN framework~\cite{Ren:2015aa} (FRCNN) for fruit detection and achieved impressive results.
Furthermore, they demonstrated that such an approach could be rapidly trained and deployed.
However, an aspect they did not consider was the potential for the system to perform not only fruit detection but also quality estimation.

Quantity and quality of fruit in a field are two key factors for determining when to harvest.
If the farmer knows there is sufficient fruit (quantity) of sufficient quality then they can allocate the workforce to harvest.
This is different from yield estimation, such as the system proposed by Nuske et al., as it requires the system to accurately estimate the number and quality of fruit present.
It has implications, not only for the farmer knowing when to deploy their labour but also for enabling automated systems, such as Harvey, to pick the most appropriate fruit, and to avoid significant losses of harvesting fruit too early.

Inspired by DeepFruits, we present a system to perform in-field assessment of fruit quantity and quality. 
Two advances are critical to this.
First, we extend the system to jointly detect the presence of a crop and estimate its quality.
To enable this we consider two network architectures that treat joint detection and quality estimation as (i) a multi-class detection problem or (ii) parallel tasks (layers).
Second, we integrate tracking into the vision system to ensure that each fruit is only counted once.
To do this we employ a tracking-via-detection approach, this allows us to use just the video stream to effectively and accurately track the crop through a sequence.



\section{Related Work}

Several researchers have considered the issue of crop detection, with one of the earliest approaches being for grapes.
Nuske et al.~\cite{Nuske_2011_6891,Nuske:2014aa} presented one of the earliest crop detection systems which was also used to estimate yield. 
Their system was developed to detect grapes using a radial symmetry transform.
The detected grapes were then used to perform accurate yield estimation. 
A limitation of their system was that they could not detect partially occluded grapes (crops). 
However, as they were performing yield estimation and not accurate crop detection they were able to cope with this limitation.

Wang et al.~\cite{Wang_2012_7240} developed an apple detection system so that they could then predict the yield.
Their detection system was based on the colour of the fruit as well as its distinctive specular reflection pattern. 
Additional information, including the average size of apples, was used to either split regions or remove erroneous detections.
A further heuristic employed, was to accept as detections only those regions which were considered mostly round. 
The detection result was then correlated to estimate the final yield.

Hung et al.~\cite{Hung:2013aa} proposed an almond segmentation approach in order to perform yield estimation.
They first learnt features using a sparse auto-encoder which were then used within a conditional random field.
They achieved impressive segmentation performance but noted that occlusion presented a major challenge and did not perform object detection.

McCool et al.~\cite{McCool16_1:conference} proposed a sweet pepper segmentation and detection approach. 
They extracted a set of features including local binary pattern features~\cite{ojala2002multiresolution}, histogram of gradients~\cite{Dalal:2005aa}, HSV colour and sparse auto-encoder features (similar to Hung et al.).
These were then used as the feature vector within a conditional random field to obtain a per-pixel segmentation.
This approach achieved impressive results similar to that of humans on the same imagery, however, its detection performance was superseded by DeepFruits.

Recently, the DeepFruits approach was proposed by Sa et al.~\cite{Sa16_1}.
This approach employs the Faster RCNN (FRCNN)~\cite{Ren:2015aa} object detection framework which uses deep neural networks to jointly perform region proposal and classification of the proposed regions.
Sa et al. achieved impressive results applying this technique to multiple crops and demonstrated that such an approach could be rapidly trained and deployed.
An aspect not considered by Sa et al. was the potential for this system to perform not only fruit detection but also quality estimation.

\section{Methods}

\begin{figure*}[!ht]
    \centering
    \subfigure[]{\includegraphics[width=0.7\textwidth]{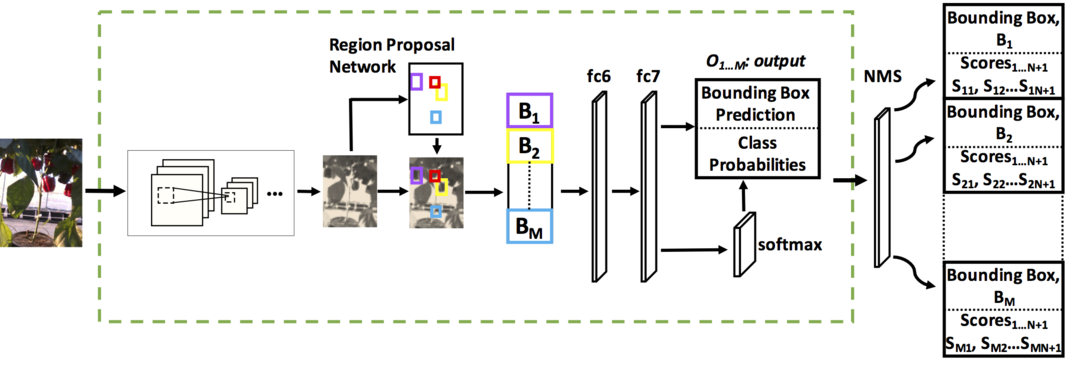}} \\
    \subfigure[]{\includegraphics[width=0.7\textwidth]{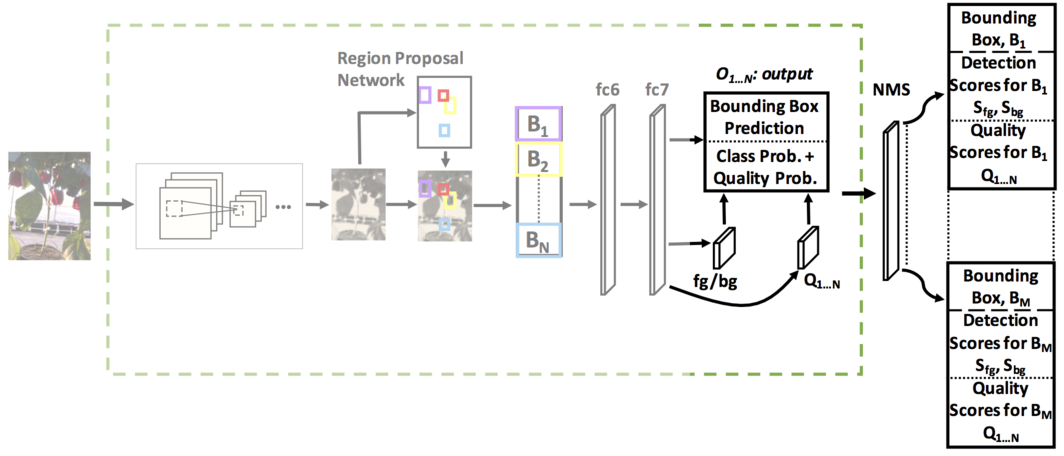}}
    \caption{In (a) is the the network structure for \textit{MultiClass-FRCNN} where there are $N$ classes representing the fruit of a particular quality and one more class for the background (bg).
    In (b) is the network structure for \textit{Parallel-FRCNN} where we highlight that the base network and region proposal is the same as before, however, instead of a single classification layer there are two parallel classification layers.}
    \label{fig:faster_rcnn_configurations}
\end{figure*}

We propose a system which locates, tracks and estimates the quality and quantity of fruit in the field.
The proposed approach consists of two sub-systems, a detection and quality estimation sub-system and a tracking sub-system.
The detection and quality estimation sub-system takes inspiration from the DeepFruit~\cite{Sa16_1} technique, but proposes a new network which jointly learns detection and quality estimation.
The tracking sub-system which calculates the quantity of fruit available for harvesting employs a tracing via detection approach.
Tracking via detection provides a simple framework to count the number of fruit (sweet pepper) that can be deployed using off-the-shelf cameras, without the need for inertia or odometry information.

\subsection{Detection and Quality Estimation}

We build on the DeepFruits approach and propose two new frameworks to jointly consider both detection and quality estimation.
The first approach is to pose the problem as a mulit-class problem where each quality is its own class that has to be detected, referred to as \textit{MultiClass-FRCNN}.
The second approach poses the problem as two parallel layers (\textit{Parallel-FRCNN}), one layer to perform detection and a second layer to perform quality estimation.
More details for the two approaches are given below.

We use a version of the FasterRCNN network implemented in TensorFlow~\cite{TF_ref}, for both of our proposals.
The FasterRCNN is a region proposal network which also outputs scores which represent the presence of desired objects or a background class. 
The network is constructed with multiple convolutional layers and finalised with fully connected layers to aggregate the filter scores.

\subsubsection{MultiClass-FRCNN: detection and quality estimation via multiple object classes}

Given $N$ quality values for the fruit we pose the detection problem as consisting of $N+1$ classes.
One class is for the negative class, or background (bg), and the remaining $N$ are positive classes, one for each of the quality values.
An illustration of this classification layer can be seen in Figure~\ref{fig:faster_rcnn_configurations} (a).

This is the simplest way to extend the DeepFruits framework to also provide quality information. 
In this formulation, each sample (positive or negative) provides a gradient from the cross-entropy loss.
A downside for this approach is that if the number of samples for a class (quality) is not balanced then learning to jointly classify and estimate the quality will likely be difficult.

\subsubsection{Parallel-FRCNN: detection and quality estimation via parallel layers}
\label{sssec:ParCNN}

In this approach, two parallel layers are introduced at the classification stage as illustrated in Figure~\ref{fig:faster_rcnn_configurations} (b).
One layer is the traditional $D+1$ class layer which differentiates the fruit, or foreground (fg), from the background; there are $D=1$ object classes (sweet pepper only in this case).
The second parallel layer is the quality estimation layer where the classes are the $N$ qualities of the fruit in question.
This layer is considered to be inactive if there is no sweet pepper (fg) present.

In terms of the backpropogated error, the \textit{detection layer} always provides a loss $L_{d}$, which is the cross-entropy loss.
The \textit{quality layer} also provides a cross-entropy loss, $L_{q}$, but only if the region in question contains the crop.
The total loss is the sum of these two losses,
\begin{equation}
    L_{tot.} = L_d + L_q.
\end{equation}
At test time, the \textit{quality layer} is only evaluated if the region is considered to contain a fruit.

\subsection{Tracking via Detection}

To ensure that we only count fruit once we track the detected fruit from the scene.
For this, we propose to use a tracking via detection approach which uses the detections, and quality estimates, provided by either the detection and quality estimation sub-system.
By using the previous sub-system, this becomes a lightweight sub-system that simply has to resolve the new detections with the active tracks.
Furthermore, as it is a vision-only system it can be deployed with cheap off-the-shelf cameras such as the Intel RealSense.

Our proposed tracking via detection approach is a two-stage technique. 
The initial frame in each image run is taken as the initialisation frame and all detected fruit within that frame are treated as unique fruit and stored as known fruit as an active \textit{track}. 
From this point forward \textit{tracks} will be used to describe the stored fruit that are being tracked and have been initialised as such, the $m$-th track is referred to as $T_{m}$.

After the initialisation stage each new frame undergoes the following procedure:
\begin{enumerate}
    \item The new detections are compared with the active tracks by calculating their intersection over union (IoU), Equation~\ref{eq:IOU}.
    \item The new detections that aren't associated with an active track are evaluated for relevance as a new fruit track by:
    \begin{enumerate}
        \item the IoU threshold, $\gamma_{merge}$, between the detection and the active tracks to determine if the detection should be considered a new track and
        \item the boundary threshold, $\gamma_{bndry}$, between the detection and the active tracks to determine if the detection should be considered a new track based on Equation~\ref{eq:bounday_condition}.
    \end{enumerate}
\end{enumerate}
These stages are described in more detail below.

In stage (1) the new detections for the $f$-th frame $\left[D_{f,1}, \dots , D_{f,K_{f}}\right]$ are compared with the $M$ active tracks $\left[T_{1}, \dots, T_{M} \right]$.
The IoU between all $M$ active tracks and all $K_{f}$ new detections is calculated and the highest matching pair, $T_{m}$ and $D_{f,k}$, is then considered.
If the The IoU between $T_{m}$ and $D_{f,k}$ is greater than a threshold, $\gamma_{dt}$, then $D_{f,k}$ is considered to be aligned with $T_{m}$.
Therefore, the position of $T_{m}$ is updated with the position $D_{k}$ and the track is considered to still be active.
The detection $D_{k}$ and active track $T_{m}$ are both removed from further consideration.
The process for stage (1) is repeated until there are no more tracks or detections to consider or when the best matching IoU is less than $\gamma_{dt}$.
If a track is not considered active for three frames then it is set to inactive and removed from the list of active tracks.

In stage (2) the remaining $J$ detections $\left[D_{f,1}, \dots , D_{f,J}\right]$ are considered as candidates to create a new track.
However, we need to resolve if the detection should be considered as a new track.
First, the detections are compared to the active tracks and their IoU is measured. 
If this IoU is above a threshold, $\gamma_{merge}$, the detection is removed form further consideration.
Second, the detections are compared to the active tracks and their boundary measure, $S_{bndry}\left(T_{m}, D_{j}\right)$ is calculated (see Equation~\ref{eq:bounday_condition}).
This boundary measure allows us to deal with cases where a small new detection is almost entirely contained within a current track, which would not be detected by the IoU.
Therefore, if the boundary measure is greater than a threshold, $\gamma_{bndry}$, the detection is removed from further consideration.
Once stage (2) has been completed for all detections, the remaining detections are used to initialise new tracks.

The boundary measure is used to cope with cases that the IoU is unable to resolve.
An example of this would be when a new small detection is almost completely contained within a large active track.
In this case, the IoU value on its own is insufficient and a new measure is required.
To deal with these cases we introduce the boundary measure which calculates how much of the $j$-th detection, $D_{j}$ is contained within the $m$-th track,
\begin{equation} \label{eq:bounday_condition}
    S_{bndry}\left(T_{m}, D_{j}\right) = \frac{\text{Area}\left(T_m\cap{D_{j}}\right)}{\text{Area}\left(D_{j}\right)}.
\end{equation}

This two-stage process allows us to update active tracks while also allowing for the addition of new tracks based on the detections.
For each track we store the following information: location (x, y locations and size of the bounding box), logit score (of the capsicum class), quality detection (based on the maximum logit colour), initialisation frame (the frame in which the track was started), and detection (if the capsicum is detected in each of the frames). 
Once a track is considered to be inactive the quality of the fruit is taken to be the most frequently occurring quality detection.

\begin{figure}[!ht]
    \centering
    \includegraphics[width=0.6\columnwidth]{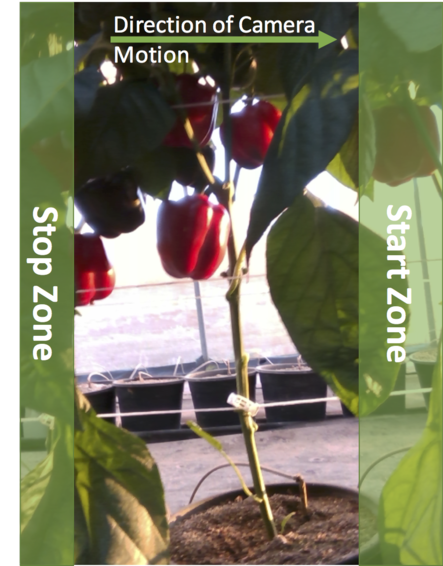}
    \caption{Example image of the start and stop zones used for tracking the objects as the objects in the image move from the right hand side to the left hand side. If the camera is moving in the opposite direction then the start and stop zones are switched.}
    \label{fig:tracking_start_stop_zone}
\end{figure}

\begin{figure}[!ht]
    \centering
    \subfigure[]{\includegraphics[width=0.3\columnwidth]{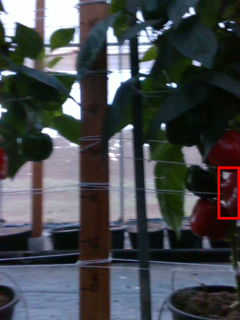}}
    \subfigure[]{\includegraphics[width=0.3\columnwidth]{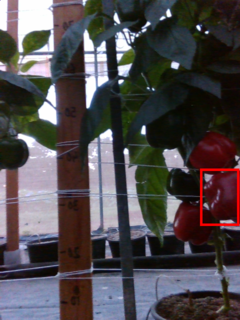}}
    \subfigure[]{\includegraphics[width=0.3\columnwidth]{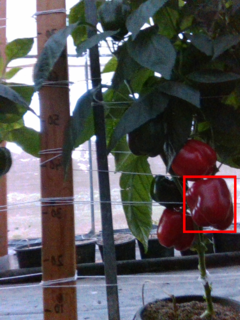}}
    \caption{Example images highlighting the difficulty in adding a detected crop whose partial presence is detected.
    In (a) the crop in the middle right is detected based on only a partial view.
    The crop becomes more visible in (b) and is fully visible in (c).}
    \label{fig:crop_edge_case}
\end{figure}

With the knowledge that the robot will be moving in one direction when employed in the field, we utilise this knowledge to introduce a final set of assumptions to further constrain the problem.
As such, we are able to define two regions that are referred to as the start and stop zones, see Figure~\ref{fig:tracking_start_stop_zone}. 
The start zone, $Z_{start}$, is a zone where we ignore new detections so that we can avoid having partial views of objects instantiate new tracks.
Using partial views to start a new track is problematic as the initial bounding box region will often be considerably smaller than later ones, see Figure~\ref{fig:crop_edge_case}.
This is difficult to resolve in a tracking via detection framework.
The stop zone, $Z_{stop}$, is used to remove detections (and consequently tracks) from further consideration before they reach the edge of the image.
Again this helps to avoid the partial view problem.




\section{Evaluation Data and Protocol}

We evaluate the proposed system on sweet pepper images obtained from a commercial farm.
Below we describe the data that was acquired, the performance measures that we use and the data splits that were used to train and evaluate the two sub-systems.

\subsection{Data Acquisition}

\begin{figure}[!t]
  \begin{center}
  \subfigure[]{\includegraphics[width=0.49\columnwidth]{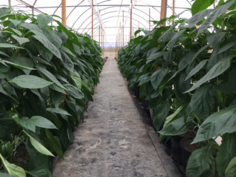}} 
  \subfigure[]{\includegraphics[width=0.49\columnwidth]{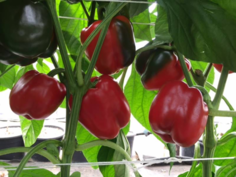}} 
  \end{center}
  \vspace{-15pt}
    \caption{
    Location of the data capture (a) is an image of protected cropping environment where the imagery of the crop (sweet pepper) was acquired.
    Example crop images are in (b) which present the green, red and mixed colour sweet peppers.}
    \label{fig:example_images}
    \vspace{-5pt}
\end{figure}

The data was acquired using an Intel RealSense SR300 with the sweet peppers (\textit{Capsicum annuum L}) grown in a commercial protective cropping structure located in Giru, North Queensland, see Figure~\ref{fig:example_images} (a).
This structure consists of a double-bay tunnel (12-m wide x 50-m long) with an arched roof (centre height: 4 m) and vertical sidewalls (height: 2.5 m).
The roof was covered with white Polyweave reinforced polyethylene film.
The sweet pepper, \textit{Red Jet} cultivar which is a blocky fruit and red when mature, were grown in a soilless media production system aligned as a single row following practices reported in Jovicich et al.~\cite{Jovicich04_1}.
The crop was arranged with plants distanced every 0.25 m and aligned in rows separated 1.45 m from each other. 

Images of the sweet pepper at varying stages of maturity were taken. 
The data acquisition was timed just prior to harvesting so there is a bias for large mature fruit.
However, there were still images of both greed, mixed colour and red sweet pepper as can be seen in Figure~\ref{fig:example_images} (b).

\begin{figure}[!t]
  \begin{center}
  \subfigure[]{\includegraphics[width=0.49\columnwidth]{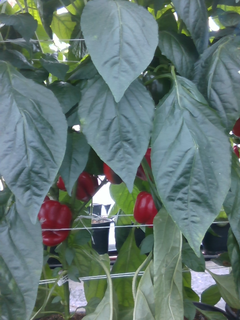}}
  \vspace{0.1cm}
  \subfigure[]{\includegraphics[width=0.49\columnwidth]{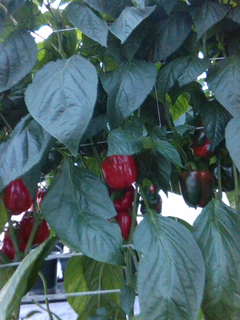}} 
  \end{center}
  \vspace{-15pt}
    \caption{
    In (a) is an image of the crop prior to leaves being removed and then (b) with the leaves removed.}
    \label{fig:crop_management_over_time}
    \vspace{-5pt}
\end{figure}

The data was acquired over three days (Tuesday, Wednesday and Thursday) and across multiple rows with leaves removed on the final day to facilitate harvesting.
Figure~\ref{fig:crop_management_over_time} (a) is an example of the crop prior to removing the leaves and Figure~\ref{fig:crop_management_over_time} (b) is an example of the crop after the leaves have been removed.
The difference between data acquired on Tuesday and Wednesday was minimal as no leaves were removed during this time.


\subsection{Evaluation Measures}

Evaluation of our systems was performed using  precision-recall curves summarised by the $F_1$ score.
For a given threshold precision (P) and recall (R) are calculated as,
\begin{align}
    \text{P}=\frac{T_P}{T_P+F_P},\;\;\text{R}=\frac{T_P}{T_P+F_N},
    \label{eq:prec_recall}
\end{align}
\noindent where $T_P$ is the number of true positives (correct detections), $F_P$ is the number of false positives (false detections), $F_N$ is the number of false negatives (miss detections), and $T_N$ is the number of true negatives (correct rejections).
The threshold chosen to calculate the $F_1$ score is the point at which the precision equals the recall,
\begin{equation}
    F_{1} = 2 \times \frac{\text{P} \cdot \text{R}}{\text{P}+\text{R}}.
    \label{eq:f1_score}
\end{equation}

We perform this analysis for varying levels of IoU for the detections.
This allows us to understand the tradeoff between how accurate the detections are and how many fruit we miss by enforcing the requirement to have a high IoU,
\begin{equation} \label{eq:IOU}
  \text{IoU} = \frac{A \cap B}{A \cup B}.
\end{equation}
where, in this case, $A$ is the groundtruth region of the fruit and $B$ is the detected region of the fruit.

\subsection{Data Splits} \label{sec:data_splits}
To train the detection and quality estimation sub-system we define a \textit{train} and \textit{eval} set.
The \textit{train} set consists of data from each of the three days, using only images captured from row 1.
The \textit{eval} set consists of data from each of the three days but only from row 3.
For both the \textit{train} and \textit{eval} sets, images are randomly selected for annotation.
This provided a range of fruit quality and density being included from the three days. 
Annotation, which includes bounding boxes and quality (green, mixed, and red) type, was completed by a single operator with checks for ambiguity from other operators.

To train the parameters of the tracking sub-system a \textit{train} and \textit{eval} set was taken from 2 days (Wednesday and Thursday) of data from row 3; the data from Tuesday was not used due to its similarity to Wednesday.
The \textit{train} set from a sub-region of the row 3 data from Thursday, for row 3 there were 4 sub-regions.
The remaining sub-regions from Thursday and all of the sub-regions from Wednesday were used as the \textit{eval} set.
Finally, to better understand the limitations of our system we included another area of the crop, across 2 days (Wednesday and Thursday), which contained juvenile plants.

\section{Results}

We present results for two sets of experiments.
First, we present results for the detection and quality estimation sub-system.
Second, using the best detection and quality estimation sub-system, we then evaluate the tracking sub-system.
In our work we use up to $N=3$ quality values for green (immature), mixed (green turning to red) and red (ripe) sweet pepper.


\subsection{Experiment I: joint detection and quality estimation}

Two approaches for joint detection and quality estimation are considered.
The first approach is to extend the number of classes so that each quality value has its own unique class, referred to as \textit{MultiClass-RCNN}.
The second approach is to have two parallel classification layers, referred to as \textit{Parallel-RCNN}, one to determine if the bounding box corresponds to a fruit and one to determine the quality of the fruit in the bounding box.

\begin{figure}
    \centering
    \subfigure[Green]{
%
%
\definecolor{mycolor1}{rgb}{1.00000,1.00000,0.00000}%
\definecolor{mycolor2}{rgb}{1.00000,0.00000,1.00000}%
\definecolor{mycolor3}{rgb}{0.00000,1.00000,1.00000}%
\begin{tikzpicture}

\begin{axis}[%
width=0.75\columnwidth,
at={(0.758in,0.481in)},
scale only axis,
xmin=0,
xmax=0.9,
xlabel={Recall Score},
ymin=0,
ymax=0.8,
ylabel={Precision Score},
axis background/.style={fill=white},
axis x line*=bottom,
axis y line*=left,
legend style={at={(0.01,0.54)},anchor=south west,legend cell align=left,align=left,draw=white!15!black},
grid=both,
grid style={line width=.1pt, draw=gray!50}
]

\addplot [color=mycolor3,solid,line width= 1.2pt]
  table[row sep=crcr]{%
0.817567567567568	0.0497124075595727\\
0.746621621621622	0.475268817204301\\
0.743243243243243	0.517647058823529\\
0.743243243243243	0.544554455445545\\
0.736486486486487	0.575197889182058\\
0.736486486486487	0.589189189189189\\
0.736486486486487	0.597260273972603\\
0.736486486486487	0.605555555555556\\
0.733108108108108	0.616477272727273\\
0.726351351351351	0.623188405797101\\
0.719594594594595	0.626470588235294\\
0.709459459459459	0.626865671641791\\
0.706081081081081	0.629518072289157\\
0.702702702702703	0.643962848297214\\
0.702702702702703	0.658227848101266\\
0.699324324324324	0.669902912621359\\
0.692567567567568	0.687919463087248\\
0.689189189189189	0.701030927835052\\
0.675675675675676	0.735294117647059\\
0.628378378378378	0.762295081967213\\
};
\addlegendentry{IOU = 0.3};

\addplot [color=red,solid,line width= 1.2pt]
  table[row sep=crcr]{%
0.763513513513513	0.0464256368118324\\
0.706081081081081	0.449462365591398\\
0.706081081081081	0.491764705882353\\
0.706081081081081	0.517326732673267\\
0.699324324324324	0.546174142480211\\
0.699324324324324	0.559459459459459\\
0.699324324324324	0.567123287671233\\
0.699324324324324	0.575\\
0.695945945945946	0.585227272727273\\
0.695945945945946	0.597101449275362\\
0.689189189189189	0.6\\
0.682432432432432	0.602985074626866\\
0.679054054054054	0.605421686746988\\
0.675675675675676	0.619195046439629\\
0.675675675675676	0.632911392405063\\
0.675675675675676	0.647249190938511\\
0.672297297297297	0.667785234899329\\
0.672297297297297	0.683848797250859\\
0.655405405405405	0.713235294117647\\
0.608108108108108	0.737704918032787\\
};
\addlegendentry{IOU = 0.4};

\addplot [color=green,solid,line width= 1.2pt]
  table[row sep=crcr]{%
0.72972972972973	0.0443714050944947\\
0.668918918918919	0.425806451612903\\
0.668918918918919	0.465882352941176\\
0.668918918918919	0.49009900990099\\
0.662162162162162	0.517150395778364\\
0.662162162162162	0.52972972972973\\
0.662162162162162	0.536986301369863\\
0.662162162162162	0.544444444444444\\
0.658783783783784	0.553977272727273\\
0.658783783783784	0.565217391304348\\
0.652027027027027	0.567647058823529\\
0.648648648648649	0.573134328358209\\
0.64527027027027	0.575301204819277\\
0.641891891891892	0.588235294117647\\
0.638513513513513	0.598101265822785\\
0.638513513513513	0.611650485436893\\
0.631756756756757	0.62751677852349\\
0.631756756756757	0.642611683848797\\
0.618243243243243	0.672794117647059\\
0.570945945945946	0.692622950819672\\
};
\addlegendentry{IOU = 0.5};

\addplot [color=blue,solid,line width= 1.2pt]
  table[row sep=crcr]{%
0.638513513513513	0.0388249794576828\\
0.597972972972973	0.380645161290323\\
0.594594594594595	0.414117647058824\\
0.594594594594595	0.435643564356436\\
0.587837837837838	0.45910290237467\\
0.587837837837838	0.47027027027027\\
0.587837837837838	0.476712328767123\\
0.587837837837838	0.483333333333333\\
0.587837837837838	0.494318181818182\\
0.587837837837838	0.504347826086957\\
0.581081081081081	0.505882352941176\\
0.577702702702703	0.51044776119403\\
0.574324324324324	0.512048192771084\\
0.570945945945946	0.523219814241486\\
0.567567567567568	0.531645569620253\\
0.567567567567568	0.54368932038835\\
0.560810810810811	0.557046979865772\\
0.560810810810811	0.570446735395189\\
0.550675675675676	0.599264705882353\\
0.516891891891892	0.627049180327869\\
};
\addlegendentry{IOU = 0.6};

\addplot [color=magenta,solid,line width= 1.2pt]
  table[row sep=crcr]{%
0.496621621621622	0.0301972062448644\\
0.462837837837838	0.294623655913978\\
0.462837837837838	0.322352941176471\\
0.462837837837838	0.339108910891089\\
0.456081081081081	0.356200527704486\\
0.456081081081081	0.364864864864865\\
0.456081081081081	0.36986301369863\\
0.456081081081081	0.375\\
0.452702702702703	0.380681818181818\\
0.452702702702703	0.388405797101449\\
0.445945945945946	0.388235294117647\\
0.445945945945946	0.394029850746269\\
0.442567567567568	0.394578313253012\\
0.439189189189189	0.402476780185759\\
0.439189189189189	0.411392405063291\\
0.439189189189189	0.420711974110032\\
0.435810810810811	0.432885906040268\\
0.435810810810811	0.443298969072165\\
0.425675675675676	0.463235294117647\\
0.408783783783784	0.495901639344262\\
};
\addlegendentry{IOU = 0.7};

\addplot [color=black,solid,line width= 1.2pt]
  table[row sep=crcr]{%
0.243243243243243	0.0147904683648316\\
0.236486486486486	0.150537634408602\\
0.236486486486486	0.164705882352941\\
0.236486486486486	0.173267326732673\\
0.233108108108108	0.182058047493404\\
0.233108108108108	0.186486486486487\\
0.233108108108108	0.189041095890411\\
0.233108108108108	0.191666666666667\\
0.233108108108108	0.196022727272727\\
0.233108108108108	0.2\\
0.226351351351351	0.197058823529412\\
0.226351351351351	0.2\\
0.222972972972973	0.198795180722892\\
0.219594594594595	0.201238390092879\\
0.219594594594595	0.205696202531646\\
0.219594594594595	0.210355987055016\\
0.219594594594595	0.218120805369128\\
0.219594594594595	0.223367697594502\\
0.212837837837838	0.231617647058824\\
0.199324324324324	0.241803278688525\\
};
\addlegendentry{IOU = 0.8};

\addplot [color=mycolor3,only marks,mark=asterisk,mark options={solid},forget plot, mark size=3pt, line width= 1.2pt]
  table[row sep=crcr]{%
0.675675675675676	0.735294117647059\\
};
\addplot [color=red,only marks,mark=asterisk,mark options={solid},forget plot, mark size=3pt, line width= 1.2pt]
  table[row sep=crcr]{%
0.655405405405405	0.713235294117647\\
};
\addplot [color=green,only marks,mark=asterisk,mark options={solid},forget plot, mark size=3pt, line width= 1.2pt]
  table[row sep=crcr]{%
0.618243243243243	0.672794117647059\\
};
\addplot [color=blue,only marks,mark=asterisk,mark options={solid},forget plot, mark size=3pt, line width= 1.2pt]
  table[row sep=crcr]{%
0.550675675675676	0.599264705882353\\
};
\addplot [color=magenta,only marks,mark=asterisk,mark options={solid},forget plot, mark size=3pt, line width= 1.2pt]
  table[row sep=crcr]{%
0.408783783783784	0.495901639344262\\
};
\addplot [color=black,only marks,mark=asterisk,mark options={solid},forget plot, mark size=3pt, line width= 1.2pt]
  table[row sep=crcr]{%
0.212837837837838	0.231617647058824\\
};
\end{axis}
\end{tikzpicture}%
} \\

    \subfigure[Mixed]{
%
%
\definecolor{mycolor1}{rgb}{1.00000,1.00000,0.00000}%
\definecolor{mycolor2}{rgb}{1.00000,0.00000,1.00000}%
\definecolor{mycolor3}{rgb}{0.00000,1.00000,1.00000}%
\begin{tikzpicture}

\begin{axis}[%
width=0.75\columnwidth,
at={(0.758in,0.481in)},
scale only axis,
xmin=0,
xmax=1,
xlabel={Recall Score},
ymin=0,
ymax=0.7,
ylabel={Precision Score},
axis background/.style={fill=white},
axis x line*=bottom,
axis y line*=left,
legend style={at={(0.01,0.54)},anchor=south west,legend cell align=left,align=left,draw=white!15!black},
grid=both,
grid style={line width=.1pt, draw=gray!50}
]

\addplot [color=mycolor3,solid,line width= 1.2pt]
  table[row sep=crcr]{%
0.970588235294118	0.0162281780181952\\
0.794117647058823	0.220408163265306\\
0.764705882352941	0.273684210526316\\
0.720588235294118	0.293413173652695\\
0.705882352941177	0.317880794701987\\
0.691176470588235	0.353383458646617\\
0.647058823529412	0.369747899159664\\
0.617647058823529	0.378378378378378\\
0.617647058823529	0.39622641509434\\
0.588235294117647	0.408163265306122\\
0.573529411764706	0.410526315789474\\
0.558823529411765	0.422222222222222\\
0.544117647058823	0.45679012345679\\
0.485294117647059	0.445945945945946\\
0.455882352941176	0.436619718309859\\
0.411764705882353	0.444444444444444\\
0.397058823529412	0.457627118644068\\
0.367647058823529	0.454545454545455\\
0.367647058823529	0.543478260869565\\
0.338235294117647	0.605263157894737\\
};
\addlegendentry{IOU = 0.3};

\addplot [color=red,solid,line width= 1.2pt]
  table[row sep=crcr]{%
0.955882352941177	0.0159822965330711\\
0.779411764705882	0.216326530612245\\
0.75	0.268421052631579\\
0.705882352941177	0.287425149700599\\
0.676470588235294	0.304635761589404\\
0.661764705882353	0.338345864661654\\
0.617647058823529	0.352941176470588\\
0.588235294117647	0.36036036036036\\
0.588235294117647	0.377358490566038\\
0.558823529411765	0.387755102040816\\
0.544117647058823	0.389473684210526\\
0.529411764705882	0.4\\
0.514705882352941	0.432098765432099\\
0.470588235294118	0.432432432432432\\
0.441176470588235	0.422535211267606\\
0.382352941176471	0.412698412698413\\
0.382352941176471	0.440677966101695\\
0.352941176470588	0.436363636363636\\
0.352941176470588	0.521739130434783\\
0.323529411764706	0.578947368421053\\
};
\addlegendentry{IOU = 0.4};

\addplot [color=green,solid,line width= 1.2pt]
  table[row sep=crcr]{%
0.911764705882353	0.0152446520776986\\
0.705882352941177	0.195918367346939\\
0.661764705882353	0.236842105263158\\
0.617647058823529	0.251497005988024\\
0.602941176470588	0.271523178807947\\
0.588235294117647	0.300751879699248\\
0.544117647058823	0.310924369747899\\
0.514705882352941	0.315315315315315\\
0.514705882352941	0.330188679245283\\
0.5	0.346938775510204\\
0.485294117647059	0.347368421052632\\
0.470588235294118	0.355555555555556\\
0.455882352941176	0.382716049382716\\
0.411764705882353	0.378378378378378\\
0.397058823529412	0.380281690140845\\
0.338235294117647	0.365079365079365\\
0.338235294117647	0.389830508474576\\
0.308823529411765	0.381818181818182\\
0.294117647058824	0.434782608695652\\
0.279411764705882	0.5\\
};
\addlegendentry{IOU = 0.5};

\addplot [color=blue,solid,line width= 1.2pt]
  table[row sep=crcr]{%
0.823529411764706	0.0137693631669535\\
0.632352941176471	0.175510204081633\\
0.617647058823529	0.221052631578947\\
0.573529411764706	0.233532934131737\\
0.558823529411765	0.251655629139073\\
0.529411764705882	0.270676691729323\\
0.485294117647059	0.277310924369748\\
0.455882352941176	0.279279279279279\\
0.455882352941176	0.292452830188679\\
0.441176470588235	0.306122448979592\\
0.426470588235294	0.305263157894737\\
0.411764705882353	0.311111111111111\\
0.397058823529412	0.333333333333333\\
0.367647058823529	0.337837837837838\\
0.352941176470588	0.338028169014085\\
0.294117647058824	0.317460317460317\\
0.294117647058824	0.338983050847458\\
0.264705882352941	0.327272727272727\\
0.264705882352941	0.391304347826087\\
0.264705882352941	0.473684210526316\\
};
\addlegendentry{IOU = 0.6};

\addplot [color=magenta,solid,line width= 1.2pt]
  table[row sep=crcr]{%
0.691176470588235	0.011556429800836\\
0.514705882352941	0.142857142857143\\
0.5	0.178947368421053\\
0.455882352941176	0.18562874251497\\
0.426470588235294	0.19205298013245\\
0.397058823529412	0.203007518796992\\
0.352941176470588	0.201680672268908\\
0.338235294117647	0.207207207207207\\
0.338235294117647	0.216981132075472\\
0.323529411764706	0.224489795918367\\
0.308823529411765	0.221052631578947\\
0.294117647058824	0.222222222222222\\
0.279411764705882	0.234567901234568\\
0.235294117647059	0.216216216216216\\
0.235294117647059	0.225352112676056\\
0.205882352941176	0.222222222222222\\
0.205882352941176	0.23728813559322\\
0.191176470588235	0.236363636363636\\
0.191176470588235	0.282608695652174\\
0.191176470588235	0.342105263157895\\
};
\addlegendentry{IOU = 0.7};

\addplot [color=black,solid,line width= 1.2pt]
  table[row sep=crcr]{%
0.426470588235294	0.00713056306860093\\
0.338235294117647	0.0938775510204082\\
0.323529411764706	0.115789473684211\\
0.308823529411765	0.125748502994012\\
0.294117647058824	0.132450331125828\\
0.294117647058824	0.150375939849624\\
0.279411764705882	0.159663865546218\\
0.279411764705882	0.171171171171171\\
0.279411764705882	0.179245283018868\\
0.264705882352941	0.183673469387755\\
0.264705882352941	0.189473684210526\\
0.25	0.188888888888889\\
0.235294117647059	0.197530864197531\\
0.191176470588235	0.175675675675676\\
0.191176470588235	0.183098591549296\\
0.176470588235294	0.19047619047619\\
0.176470588235294	0.203389830508475\\
0.161764705882353	0.2\\
0.161764705882353	0.239130434782609\\
0.161764705882353	0.289473684210526\\
};
\addlegendentry{IOU = 0.8};

\addplot [color=mycolor3,only marks,mark=asterisk,mark options={solid},forget plot, mark size=3pt, line width= 1.2pt]
  table[row sep=crcr]{%
0.544117647058823	0.45679012345679\\
};
\addplot [color=red,only marks,mark=asterisk,mark options={solid},forget plot, mark size=3pt, line width= 1.2pt]
  table[row sep=crcr]{%
0.514705882352941	0.432098765432099\\
};
\addplot [color=green,only marks,mark=asterisk,mark options={solid},forget plot, mark size=3pt, line width= 1.2pt]
  table[row sep=crcr]{%
0.455882352941176	0.382716049382716\\
};
\addplot [color=blue,only marks,mark=asterisk,mark options={solid},forget plot, mark size=3pt, line width= 1.2pt]
  table[row sep=crcr]{%
0.397058823529412	0.333333333333333\\
};
\addplot [color=magenta,only marks,mark=asterisk,mark options={solid},forget plot, mark size=3pt, line width= 1.2pt]
  table[row sep=crcr]{%
0.397058823529412	0.203007518796992\\
};
\addplot [color=black,only marks,mark=asterisk,mark options={solid},forget plot, mark size=3pt, line width= 1.2pt]
  table[row sep=crcr]{%
0.264705882352941	0.189473684210526\\
};
\end{axis}
\end{tikzpicture}%
} \\

    \subfigure[Red]{
%
%
\definecolor{mycolor1}{rgb}{1.00000,1.00000,0.00000}%
\definecolor{mycolor2}{rgb}{1.00000,0.00000,1.00000}%
\definecolor{mycolor3}{rgb}{0.00000,1.00000,1.00000}%
\begin{tikzpicture}

\begin{axis}[%
width=0.75\columnwidth,
at={(0.758in,0.481in)},
scale only axis,
xmin=0,
xmax=1,
xlabel={Recall Score},
ymin=0,
ymax=0.8,
ylabel={Precision Score},
axis background/.style={fill=white},
axis x line*=bottom,
axis y line*=left,
legend style={at={(0.01,0.54)},anchor=south west,legend cell align=left,align=left,draw=white!15!black},
grid=both,
grid style={line width=.1pt, draw=gray!50}
]

\addplot [color=mycolor3,solid,line width= 1.2pt]
  table[row sep=crcr]{%
0.909448818897638	0.104007203962179\\
0.87992125984252	0.480645161290323\\
0.87007874015748	0.507462686567164\\
0.868110236220472	0.533252720677146\\
0.868110236220472	0.547146401985112\\
0.864173228346457	0.559948979591837\\
0.860236220472441	0.565329883570505\\
0.852362204724409	0.572751322751323\\
0.850393700787402	0.580645161290323\\
0.848425196850394	0.59122085048011\\
0.848425196850394	0.596952908587258\\
0.84251968503937	0.605374823196605\\
0.832677165354331	0.613933236574746\\
0.826771653543307	0.626865671641791\\
0.824803149606299	0.645608628659476\\
0.818897637795276	0.652037617554859\\
0.812992125984252	0.668284789644013\\
0.803149606299213	0.678868552412646\\
0.791338582677165	0.696707105719237\\
0.769685039370079	0.728119180633147\\
};
\addlegendentry{IOU = 0.3};

\addplot [color=red,solid,line width= 1.2pt]
  table[row sep=crcr]{%
0.87992125984252	0.10063034669068\\
0.848425196850394	0.463440860215054\\
0.844488188976378	0.492537313432836\\
0.838582677165354	0.515114873035067\\
0.834645669291339	0.52605459057072\\
0.830708661417323	0.538265306122449\\
0.826771653543307	0.543337645536869\\
0.818897637795276	0.55026455026455\\
0.81496062992126	0.556451612903226\\
0.812992125984252	0.566529492455418\\
0.812992125984252	0.57202216066482\\
0.809055118110236	0.581329561527581\\
0.80511811023622	0.593613933236575\\
0.797244094488189	0.604477611940298\\
0.795275590551181	0.622496147919877\\
0.789370078740158	0.628526645768025\\
0.78740157480315	0.647249190938511\\
0.781496062992126	0.660565723793677\\
0.765748031496063	0.674176776429809\\
0.746062992125984	0.705772811918063\\
};
\addlegendentry{IOU = 0.4};

\addplot [color=green,solid,line width= 1.2pt]
  table[row sep=crcr]{%
0.830708661417323	0.095002251238181\\
0.793307086614173	0.433333333333333\\
0.789370078740158	0.460390355912744\\
0.785433070866142	0.482466747279323\\
0.785433070866142	0.495037220843672\\
0.781496062992126	0.506377551020408\\
0.779527559055118	0.51228978007762\\
0.769685039370079	0.517195767195767\\
0.763779527559055	0.521505376344086\\
0.759842519685039	0.529492455418381\\
0.759842519685039	0.534626038781163\\
0.753937007874016	0.541725601131542\\
0.748031496062992	0.551523947750363\\
0.740157480314961	0.561194029850746\\
0.738188976377953	0.577812018489985\\
0.734251968503937	0.584639498432602\\
0.734251968503937	0.603559870550162\\
0.726377952755906	0.613976705490849\\
0.714566929133858	0.629116117850953\\
0.698818897637795	0.661080074487896\\
};
\addlegendentry{IOU = 0.5};

\addplot [color=blue,solid,line width= 1.2pt]
  table[row sep=crcr]{%
0.759842519685039	0.0868977937865826\\
0.728346456692913	0.397849462365591\\
0.724409448818898	0.422502870264064\\
0.718503937007874	0.441354292623942\\
0.718503937007874	0.452853598014888\\
0.716535433070866	0.464285714285714\\
0.71259842519685	0.468305304010349\\
0.704724409448819	0.473544973544974\\
0.700787401574803	0.478494623655914\\
0.69488188976378	0.484224965706447\\
0.692913385826772	0.487534626038781\\
0.688976377952756	0.495049504950495\\
0.68503937007874	0.505079825834543\\
0.679133858267717	0.514925373134328\\
0.677165354330709	0.530046224961479\\
0.671259842519685	0.53448275862069\\
0.671259842519685	0.551779935275081\\
0.663385826771654	0.560732113144759\\
0.655511811023622	0.577123050259965\\
0.633858267716535	0.599627560521415\\
};
\addlegendentry{IOU = 0.6};

\addplot [color=magenta,solid,line width= 1.2pt]
  table[row sep=crcr]{%
0.622047244094488	0.0711391265195858\\
0.590551181102362	0.32258064516129\\
0.586614173228346	0.342135476463835\\
0.582677165354331	0.357920193470375\\
0.580708661417323	0.366004962779156\\
0.578740157480315	0.375\\
0.574803149606299	0.377749029754204\\
0.568897637795276	0.382275132275132\\
0.568897637795276	0.388440860215054\\
0.56496062992126	0.393689986282579\\
0.562992125984252	0.39612188365651\\
0.562992125984252	0.404526166902405\\
0.561023622047244	0.413642960812772\\
0.557086614173228	0.422388059701493\\
0.55511811023622	0.434514637904468\\
0.551181102362205	0.438871473354232\\
0.551181102362205	0.453074433656958\\
0.549212598425197	0.464226289517471\\
0.539370078740158	0.474870017331023\\
0.523622047244094	0.495344506517691\\
};
\addlegendentry{IOU = 0.7};

\addplot [color=black,solid, line width= 1.2pt]
  table[row sep=crcr]{%
0.387795275590551	0.0443493921656911\\
0.375984251968504	0.205376344086022\\
0.375984251968504	0.219288174512055\\
0.375984251968504	0.230955259975816\\
0.375984251968504	0.23697270471464\\
0.372047244094488	0.241071428571429\\
0.37007874015748	0.243208279430789\\
0.366141732283465	0.246031746031746\\
0.366141732283465	0.25\\
0.366141732283465	0.255144032921811\\
0.366141732283465	0.257617728531856\\
0.366141732283465	0.263083451202263\\
0.364173228346457	0.268505079825835\\
0.362204724409449	0.274626865671642\\
0.360236220472441	0.281972265023112\\
0.356299212598425	0.283699059561129\\
0.356299212598425	0.292880258899676\\
0.356299212598425	0.301164725457571\\
0.350393700787402	0.308492201039861\\
0.34251968503937	0.324022346368715\\
};
\addlegendentry{IOU = 0.8};

\addplot [color=mycolor3,only marks,mark=asterisk,mark options={solid},forget plot, mark size=3pt, line width= 1.2pt]
  table[row sep=crcr]{%
0.769685039370079	0.728119180633147\\
};
\addplot [color=red,only marks,mark=asterisk,mark options={solid},forget plot, mark size=3pt, line width= 1.2pt]
  table[row sep=crcr]{%
0.746062992125984	0.705772811918063\\
};
\addplot [color=green,only marks,mark=asterisk,mark options={solid},forget plot, mark size=3pt, line width= 1.2pt]
  table[row sep=crcr]{%
0.698818897637795	0.661080074487896\\
};
\addplot [color=blue,only marks,mark=asterisk,mark options={solid},forget plot, mark size=3pt, line width= 1.2pt]
  table[row sep=crcr]{%
0.633858267716535	0.599627560521415\\
};
\addplot [color=magenta,only marks,mark=asterisk,mark options={solid},forget plot, mark size=3pt, line width= 1.2pt]
  table[row sep=crcr]{%
0.523622047244094	0.495344506517691\\
};
\addplot [color=black,only marks,mark=asterisk,mark options={solid},forget plot, mark size=3pt, line width= 1.2pt]
  table[row sep=crcr]{%
0.34251968503937	0.324022346368715\\
};
\end{axis}
\end{tikzpicture}%
} 
%
    


    \caption{Precision-recall curves for the \textit{MultiClass-RCNN} approach for (a) green, (b) mixed and (c) red fruit.}
    \label{fig:mclass_detections}
\end{figure}
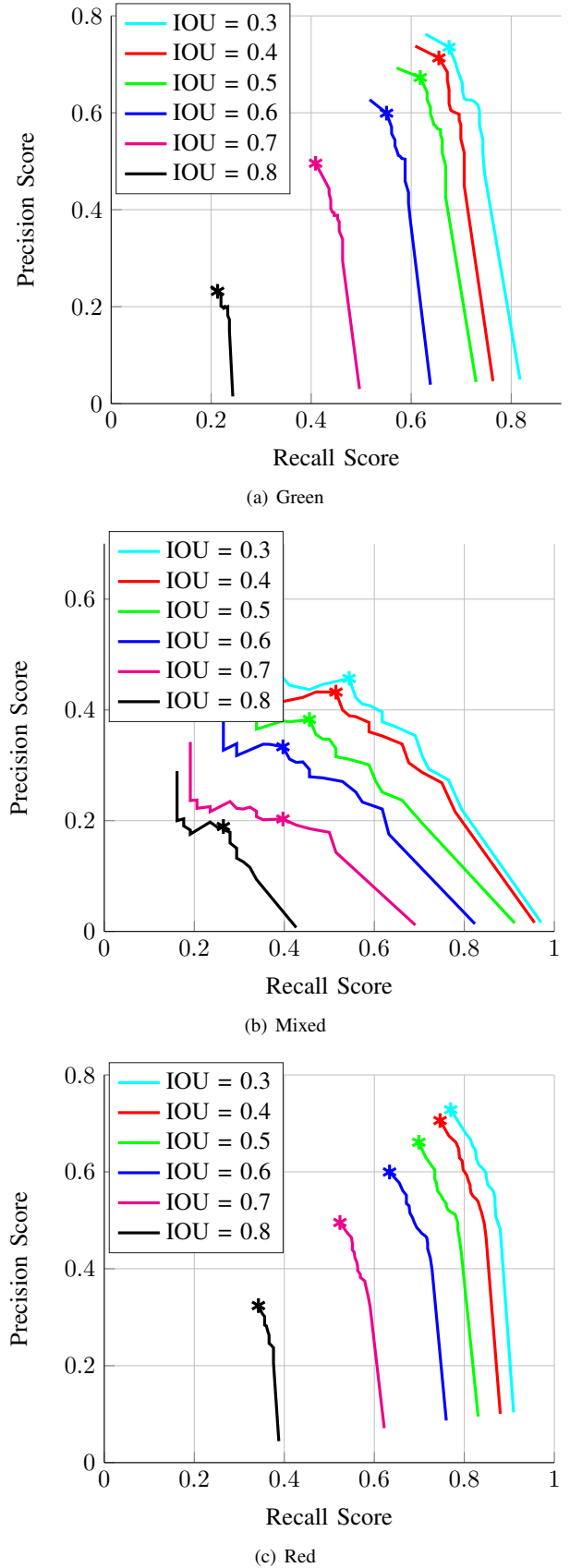

The detection performance of the \textit{MultiClass-RCNN} is class dependent.
This can be seen in Figure~\ref{fig:mclass_detections} where the performance of the $3$ classes varies considerably, with the mixed class having the lowest performance with an $F_1$ score of $46.9$ at an IoU of $0.4$\footnote{An IoU of $0.4$ was chosen as it provided high accuracy and good quality bounding box detections.}.
Detection of green and red fruit is considerably higher with $F_1=68.3$ and $F_1=72.5$ respectively, at $IoU=0.4$.


Analysis of the poor performance of the \textit{MultiClass-RCNN} system in detecting the mixed fruit can be attributed to the limited training data for this class.
From 152 images, both green and red fruit have a large number of training samples with 253 and 636 respectively.
By comparison, the mixed fruit has just 69 training samples.
The testing sample sizes remain consistent in distribution with the training samples, where in 133 images we aim to classify 296 green, 508 red, and only 68 mixed fruit.
We believe that the disparity in these training sample sizes results in a poor classifier being trained to detect the mixed fruit. 
By contrast, the technique employed in the \textit{Parallel-RCNN} provides a mechanism to nullify this problem.

\begin{figure}[htb!]
    \centering
    \input{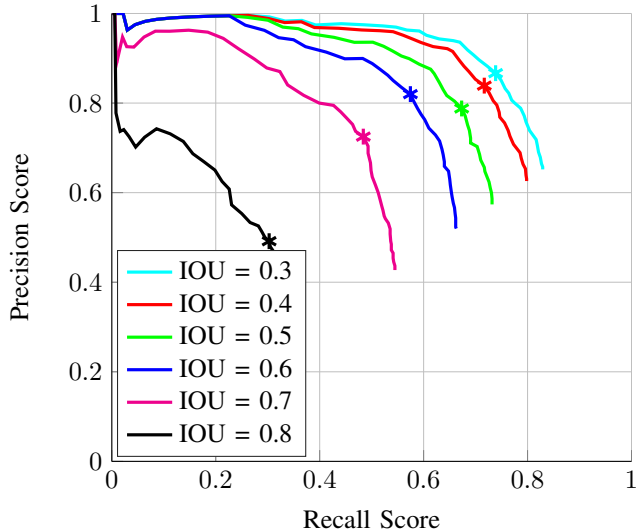}
    \caption{Precision recall scores of the parallel class classifier against varying levels of IOU threshold. Scores are only considered if the capsicum class logit score is returned as the highest score (i.e. if background returns a higher logit score the regions is disregarded).}
    \label{fig:parallel_classifier}
\end{figure}

The detection performance of the \textit{Parallel-RCNN} for any fruit (green, mixed or red) is superior to the best performing class for \textit{MultiClass-RCNN}.
At IoU=0.4 the \textit{Parallel-RCNN} has an $F_{1} = 77.3$ for detecting any fruit, this is superior to the $F_{1} = 72.5$ for the best performing class (red) of the \textit{MultiClass-RCNN}.
The full results for this system are outlined in Figure~\ref{fig:parallel_classifier}, where the superior performance of the \textit{Parallel-RCNN} technique is evident at all IoU values.

\begin{table}[!t]
\centering 
\caption{The confusion matrix for classifying the quality via overall colour of the crop, IoU=0.4} 
\begin{tabular}{| l | c | c | c |}  
\cline{2-4} 
\multicolumn{1}{l|}{}          & Green     & Mixed   & Red       \\ \hline
Green       & 94.0\%    & 4.8\%   & 1.2\%     \\ \hline
Mixed       & 13.6\%    & 62.7\%  & 23.7\%    \\ \hline
Red         & 0.7\%     & 9.8\%   & 89.5\%    \\ \hline
\end{tabular}
\label{tbl:parallel_quality_perf}
\end{table}

The quality estimation performance (green, mixed, red) of the \textit{Parallel-RCNN} is summarised by the confusion matrix in Table~\ref{tbl:parallel_quality_perf}.
On average, the quality performance of the system is $82.1\%$. 
As the highest performing quality factor, green fruit achieves an accuracy of $94.0\%$, also performing strongly the red fruit is correctly classified $89.5\%$ of the time.
However, the performance of the mixed (green turning to red) fruit is considerably lower at $62.7\%$ which is most often confused with the red fruit, $23.7\%$ of the time.
This highlights the challenge of correctly classifying the mixed fruit from images collected in the field.

%
%



\subsection{Experiment II: tracking-via-detection}

The tracking via detection sub-system takes the detection and quality proposal from the \textit{Parallel-RCNN} sub-system.
Formulating the approach in this manner allows for the aggregation of the number, and quality, of the fruit in the field.
After empirical evaluation, for our experiments we set the size of the start and stop zones to $Z_{start}=0.2$ and $Z_{stop}=0.15$ respectively. 
The parameters of the tracker ($\gamma_{dt}$, $\gamma_{merge}$ and $\gamma_{boundary}$) were optimised on its associated \textit{train} set, see Section~\ref{sec:data_splits}.
The optimal values for each of these was found to be $\gamma_{dt}=0.3$, $\gamma_{merge}=0.4$ and $\gamma_{boundary}=0.5$.
Using these parameters we then evaluated the performance of the tracker on the \textit{eval} set.

The performance of the tracking system was evaluated over two days (Wednesday and Thursday). 
Table~\ref{tab:main_results} presents a summary of these results and demonstrates the potential of this approach.
Across the two days the system is able to predict the total number of fruit to within 4.1\% of the ground truth (GT).
For Wednesday and Thursday the system is able to predict the total number of fruit to within 11.8\% and 4.9\% of the ground truth (GT) respectively.
We attribute the lower performance on Wednesday to the more challenging conditions provided by the presence of the foliage.

\begin{table}[ht]
    \centering
    \caption{Tracking results for Wednesday and Thursday.}
    \begin{tabular}{|c|c|c|c|c|}
        \hline
            Day                   & Green & Mixed & Red  & Total \\
        \hline
            Thursday (Estimated)  & 95    & 17    & 103   & 215 \\
            Thursday (GT)         & 84    & 15    & 106  & 205 \\
        \hline
            Wednesday (Estimated) & 64    & 21    & 123  & 209 \\
            Wednesday (GT)        & 95    & 19    & 124  & 237 \\
        \hline    
            Combined (Estimated)  & 159   & 38    & 227 & 424 \\
            Combined (GT)         & 179   & 34    & 229 & 442 \\
        \hline 
            Combined Percent Error  & 11.2\% & 11.8\% & 0.9\% & 4.1\% \\
        \hline
    \end{tabular}
    \label{tab:main_results}
\end{table}

Further examination of the performance for fruit quality indicates that the system is accurate.
Across the two days the system estimates the number of red, green and mixed crop within 0.9\%, 11.2\% and 11.8\% respectively.
The main errors for the estimate of green fruit occurs on Wednesday where 64 fruit are found out of the 95 which are present.
We partially attribute this issue to the dense foliage present for Wednesday.
Finally, to better understand the limits of the system we apply it to another section of the crop which contains only green fruit.

In Table~\ref{tab:green_capsicum} we present the performance of the tracking system for green fruit which is a mixture of juvenile and larger fruit.
It can be seen that the performance for this area of the crop is considerably lower than for the larger fruit.
This low performance occurs because the detector is not detecting the juvenile fruit, for which it was not trained for.
We also note that the system correctly identified that there were no mixed or red colour fruit present, further highlighting the robustness of the system.

\begin{table}[ht]
    \centering
    \caption{Tracking results for Thursday (Sec. 5) and Wednesday (Sec. 5) for juvenile sweet pepper.}
    \begin{tabular}{|c|c|c|c|c|}
        \hline
            Day                   & Green & Mixed & Red  & Total \\
        \hline
            Thursday (Estimated)  & 29    & 0     & 0    & 29 \\
            Thursday (GT)         & 44    & 0     & 0    & 44 \\
        \hline
            Wednesday (Estimated) & 22    & 0     & 0    & 22 \\
            Wednesday (GT)        & 43    & 0     & 0    & 43 \\
        \hline    
            Combined (Estimated)  & 51   & 0    & 0 & 51 \\
            Combined (GT)         & 87   & 0    & 0 & 87 \\
        \hline
    \end{tabular}
    \label{tab:green_capsicum}
\end{table}

\balance 

\section{Conclusion}

We have presented a vision-only system that can accurately estimate the quantity and quality of sweet pepper, a key horticultural crop. 
Empirically, we have demonstrated that joint detection and quality estimation can be performed using a \textit{Parallel-RFCNN} structure.
Such an approach can accurately detect fruit with $F_{1}=77.3$, at an $IoU=0.4$, and can accurately estimate its quality with an average accuracy of 82.1\%.

A tracking via detection system is then employed to accurately count the fruit present in the field.
This tracking approach is a vision-only solution and as such is cheap to implement as it only requires a camera.
In experiments across 2 days we show that our proposed system can accurately estimate the number of sweet pepper present to within 4.1\% of the ground truth.

One limitation witnessed in the tracking system was the reliance on the accuracy of the detection technique.
For instance, our detection system is trained to detect crop as they near maturity.
We show that when this is applied to early stage fruit, which are visually very small, this leads to errors in the estimate of the number of fruit present.
In this specific case the technique was able to predicting the presence of 57 out of the 87 sweet pepper present in the field.

Despite experiencing limitations within the proposal, we present a new technique to the horticulture field which is able to accurately calculate the quality and quantity of sweet peppers in the field.
Our proposed approach has shown considerable ability across various conditions caused by image quality, variable lighting, foliage presence, and juvenile fruit.

\section*{ACKNOWLEDGMENT}
The authors would like to acknowledge Elio Jovicich and Heidi Wiggenhauser from Queensland Department of Agriculture and Fisheries for their advice and support. This work was funded by Hort Innovation, using the vegetable research and development levy and contributions from the Australian Government project (VG15024). Hort Innovation is the grower-owned, not-for-profit research and development corporation for Australian horticulture.

\vspace{-2mm}
\bibliographystyle{IEEEtran}
\bibliography{./bibs/mccool_bibs}

\begin{thebibliography}{10}
\providecommand{\url}[1]{#1}
\csname url@samestyle\endcsname
\providecommand{\newblock}{\relax}
\providecommand{\bibinfo}[2]{#2}
\providecommand{\BIBentrySTDinterwordspacing}{\spaceskip=0pt\relax}
\providecommand{\BIBentryALTinterwordstretchfactor}{4}
\providecommand{\BIBentryALTinterwordspacing}{\spaceskip=\fontdimen2\font plus
\BIBentryALTinterwordstretchfactor\fontdimen3\font minus
  \fontdimen4\font\relax}
\providecommand{\BIBforeignlanguage}[2]{{%
\expandafter\ifx\csname l@#1\endcsname\relax
\typeout{** WARNING: IEEEtran.bst: No hyphenation pattern has been}%
\typeout{** loaded for the language `#1'. Using the pattern for}%
\typeout{** the default language instead.}%
\else
\language=\csname l@#1\endcsname
\fi
#2}}
\providecommand{\BIBdecl}{\relax}
\BIBdecl

\bibitem{Bawden17_1}
O.~Bawden, J.~Kulk, R.~Russel, C.~McCool, and T.~Perez, ``Broadacre weed
  management robot,'' \emph{Journal of Field Robotics}, 2017.

\bibitem{Lehnert17_1}
C.~Lehnert, A.~English, C.~McCool, A.~Tow, and T.~Perez, ``Autonomous sweet
  pepper harvesting for protected cropping systems,'' \emph{IEEE Robotics and
  Automation Letters}, 2017.

\bibitem{Nuske_2011_6891}
S.~T. Nuske, S.~Achar, T.~Bates, S.~G. Narasimhan, and S.~Singh, ``Yield
  estimation in vineyards by visual grape detection,'' in \emph{Proceedings of
  the 2011 IEEE/RSJ International Conference on Intelligent Robots and Systems
  (IROS '11)}, September 2011.

\bibitem{Hung:2013aa}
C.~Hung, J.~Nieto, Z.~Taylor, J.~Underwood, and S.~Sukkarieh, ``Orchard fruit
  segmentation using multi-spectral feature learning,'' in \emph{Intelligent
  Robots and Systems (IROS), 2013 IEEE/RSJ International Conference on}, Nov
  2013, pp. 5314--5320.

\bibitem{McCool16_1:conference}
C.McCool, I.~Sa, F.~Dayoub, C.~Lehnert, T.~Perez, and B.~Upcroft, ``Visual
  detection of occluded crop: For automated harvesting,'' in \emph{IEEE
  International Conference on Robotics and Automation (ICRA)}, 2016.

\bibitem{Sa16_1}
I.~Sa, Z.~Ge, F.~Dayoub, B.~Upcroft, T.~Perez, and C.~McCool, ``Deepfruits: A
  fruit detection system using deep neural networks,'' \emph{Sensors}, 2016.

\bibitem{Ren:2015aa}
S.~Ren, K.~He, R.~Girshick, and J.~Sun, ``{Faster R-CNN: Towards real-time
  object detection with region proposal networks},'' in \emph{Advances in
  Neural Information Processing Systems}, 2015, pp. 91--99.

\bibitem{Nuske:2014aa}
S.~Nuske, K.~Wilshusen, S.~Achar, L.~Yoder, S.~Narasimhan, and S.~Singh,
  ``Automated visual yield estimation in vineyards,'' \emph{Journal of Field
  Robotics}, vol.~31, no.~5, pp. 837--860, 2014.

\bibitem{Wang_2012_7240}
Q.~Wang, S.~T. Nuske, M.~Bergerman, and S.~Singh, ``Automated crop yield
  estimation for apple orchards,'' in \emph{13th Internation Symposium on
  Experimental Robotics (ISER 2012)}, no. CMU-RI-TR-, July 2012.

\bibitem{ojala2002multiresolution}
T.~Ojala, M.~Pietikainen, and T.~Maenpaa, ``Multiresolution gray-scale and
  rotation invariant texture classification with local binary patterns,''
  \emph{IEEE Transactions on Pattern Analysis and Machine Intelligence},
  vol.~24, no.~7, pp. 971--987, 2002.

\bibitem{Dalal:2005aa}
N.~Dalal and B.~Triggs, ``Histograms of oriented gradients for human
  detection,'' in \emph{IEEE Computer Society Conference on Computer Vision and
  Pattern Recognition}, vol.~1, June 2005, pp. 886--893 vol. 1.

\bibitem{TF_ref}
\BIBentryALTinterwordspacing
M.~Abadi, A.~Agarwal, P.~Barham, E.~Brevdo, Z.~Chen, C.~Citro, G.~S. Corrado,
  A.~Davis, J.~Dean, M.~Devin, S.~Ghemawat, I.~Goodfellow, A.~Harp, G.~Irving,
  M.~Isard, Y.~Jia, R.~Jozefowicz, L.~Kaiser, M.~Kudlur, J.~Levenberg,
  D.~Man\'{e}, R.~Monga, S.~Moore, D.~Murray, C.~Olah, M.~Schuster, J.~Shlens,
  B.~Steiner, I.~Sutskever, K.~Talwar, P.~Tucker, V.~Vanhoucke, V.~Vasudevan,
  F.~Vi\'{e}gas, O.~Vinyals, P.~Warden, M.~Wattenberg, M.~Wicke, Y.~Yu, and
  X.~Zheng, ``{TensorFlow}: Large-scale machine learning on heterogeneous
  systems,'' 2015, software available from tensorflow.org. [Online]. Available:
  \url{https://www.tensorflow.org/}
\BIBentrySTDinterwordspacing

\bibitem{Jovicich04_1}
E.~Jovicich, D.~Cantliffe, and P.~Stoffella, ``Fruit yield and quality
  greenhouse-grown bell pepper as influenced by density, container, and trellis
  system,'' \emph{HortTechnology}, pp. 507--513, 2004.

\end{thebibliography}

\clearpage



%


%

\end{document}